\documentclass[wcp]{jmlr}

\usepackage{longtable}
\usepackage{multirow}
\usepackage{caption}
\usepackage{lipsum}
\usepackage{verbatim}
\usepackage{tcolorbox}
\usepackage{booktabs}

\usepackage{lineno}

\makeatletter
\let\Ginclude@graphics\@org@Ginclude@graphics 
\makeatother

\title[Short Title]{Large Vision-Language Models as Emotion Recognizers in Context Awareness}

 \author{\Name{Yuxuan Lei$^{1,2}$} \Email{yxlei22@m.fudan.edu.cn}\\
   \Name{Dingkang Yang$^{1,2,*}$} \Email{dkyang20@fudan.edu.cn}\\
   \Name{Zhaoyu Chen$^{1}$} \Email{zhaoyuchen20@fudan.edu.cn}\\
   \Name{Jiawei Chen$^{1,2}$} \Email{22210860033@m.fudan.edu.cn}\\
   \Name{Peng Zhai$^{1,2}$} \Email{pzhai@fudan.edu.cn}\\
   \Name{Lihua Zhang$^{1,2,3,4,*}$} \Email{lihuazhang@fudan.edu.cn}\\
   \addr $^{1}$ Academy for Engineering and Technology, Fudan University\\
   \addr $^{2}$ Cognition and Intelligent Technology Laboratory (CIT Lab)\\
   \addr $^{3}$ Engineering Research Center of AI and Robotics, Ministry of Education, Shanghai, China \\
   \addr $^{4}$ AI and Unmanned Systems Engineering Research Center of Jilin Province, Changchun, China
   }

\begin{document}

\maketitle

\begin{abstract}
Context-aware emotion recognition (CAER) is a complex and significant task that requires perceiving emotions from various contextual cues. Previous approaches primarily focus on designing sophisticated architectures to extract emotional cues from images. However, their knowledge is confined to specific training datasets and may reflect the subjective emotional biases of the annotators. Furthermore, acquiring large amounts of labeled data is often challenging in real-world applications. In this paper, we systematically explore the potential of leveraging Large Vision-Language Models (LVLMs) to empower the CAER task from three paradigms: 1) We fine-tune LVLMs on two CAER datasets, which is the most common way to transfer large models to downstream tasks. 2) We design zero-shot and few-shot patterns to evaluate the performance of LVLMs in scenarios with limited data or even completely unseen. 
In this case, a training-free framework is proposed to fully exploit the In-Context Learning (ICL) capabilities of LVLMs.
Specifically, we develop an image similarity-based ranking algorithm to retrieve examples; subsequently, the instructions, retrieved examples, and the test example are combined to feed  LVLMs to obtain the corresponding sentiment judgment.
3) To leverage the rich knowledge base of LVLMs, we incorporate Chain-of-Thought (CoT) into our framework to enhance the model's reasoning ability and provide interpretable results. Extensive experiments and analyses demonstrate that LVLMs achieve competitive performance in the CAER task across different paradigms. Notably, the superior performance in few-shot settings indicates the feasibility of LVLMs for accomplishing specific tasks without extensive training.
\end{abstract}
\begin{keywords}
Emotion Recognition, Context understanding, Large Vision-Language Models,  In-Context Learning
\end{keywords}

\section{Introduction}
As a crucial factor in mutual understanding, friendly communication, and maintaining long-term relationships, emotions play a vital role in our daily lives. In recent years, emotion recognition tasks have garnered increasing attention. Researchers are striving to identify and analyze the complex emotions expressed by humans from various modalities such as vision, language, and speech. So far, this technology has been applied in numerous fields (\cite{pepa2021automatic,yang2023aide}).
In the past, research on visual emotion analysis has primarily concentrated on facial expressions~\cite{jiang2020dfew}, with most facial recognition datasets providing cropped facial images. However, this does not fully align with the way humans perceive emotions through visual cues. Typically, in natural scenes, we do not only observe a person's facial expressions but also their body language (such as gestures and postures), the environment, and their interactions with others. These elements collectively contribute to our assessment of emotional states.~\cite{kosti2019context} recognized this aspect and introduced the task of Context-Aware Emotion Recognition (CAER). They combined contextual and environmental factors in static images to construct and release the EMOTIC dataset, making a significant contribution to the field of CAER. Due to the complexity and subjectivity of human emotions, fine-grained emotion recognition is challenging even for human experts. Different individuals may interpret the emotional context of the same image differently. Therefore, it can be said that there is no exact correct answer in emotion recognition. 
Traditional CAER models (\cite{kosti2017emotion,lee2019context,zhang2019context}) are typically trained and tested on fixed datasets. It raises issues of generalization, as the knowledge capacity of traditionally trained models is limited to the specific datasets they are trained on. Additionally, CAER models trained on particular datasets may learn the annotators' subjective emotional bias biases, which could cause confusion for others. Second, in real-world scenarios, it is not always feasible to obtain large amounts of labeled data. When a model is transferred to an unseen sentiment domain, collecting substantial data and retraining the model incur significant costs.

Large Language Models (LLMs) (\cite{touvron2023llama,achiam2023gpt}) have learned human thought processes from billions of data points. Their massive parameter size translates to a vast knowledge capacity, enabling them to generalize well to different downstream tasks even in few-shot or zero-shot scenarios. Large Vision-Language Models (LVLMs) (\cite{liu2024visual,lin2024vila}) combine the capabilities of LLMs with visual understanding, allowing the powerful abilities of these models to be transferred to the visual domain. By learning the correlation between vision and language from extensive image-text data, LVLMs have achieved remarkable results in tasks such as image classification (\cite{radford2021learning}), object segmentation (\cite{zhang2024makes}), and visual question answering (VQA)(\cite{alayrac2022flamingo}). However, as far as we know, the exploration of LVLMs in the CAER task remains relatively underdeveloped.

This paper aims to explore the potential of leveraging various paradigms of LVLMs to enhance the CAER task. Specifically, we first perform supervised fine-tuning (SFT) on the LVLMs, which is the most common way for generalizing large models to downstream tasks. However, in real-world scenarios, it is often challenging to obtain large amounts of labeled data, and the knowledge distribution of fine-tuned models may be disrupted. Therefore, we extend our investigation to include zero-shot and few-shot paradigms. For the few-shot setting, we aim to leverage the In-Context Learning (ICL) (\cite{brown2020language}) capability of LVLMs to achieve emotion recognition with small samples. This paradigm allows LVLMs to improve their emotion recognition performance by learning from a few examples within the context, without updating the model parameters. During the few-shot inference, LVLMs do not rely on learning the distribution of the dataset to make predictions. Instead, it generates decisions by learning from relevant tasks and leveraging its own extensive knowledge, which avoids the impact of annotation biases inherent in the dataset. It has been provided that downstream performance is highly sensitive to the choice of in-context demonstrations by ~\cite{zhang2024makes}, and a good in-context demonstration should be semantically similar to the query. To achieve this, we propose a training-free framework to fully exploit the ICL capability of LVLMs. Within our framework, we design a demonstration retrieval module based on different contexts to retrieve the top-k examples most similar to the test sample. This allows us to fully consider various aspects of the images when selecting example samples. Then, we construct a prompt template that packages the top-k selected examples ranked by similarity into the template. Finally, we concatenate the instructions, demonstrations, and target test sample to serve as input to the LVLMs and obtain the corresponding sentiment judgment.

Furthermore, to fully leverage the internal knowledge of large models, we extend the proposed framework to a few-shot Chain-of-Thought (CoT) paradigm. We utilize GPT-4 Vision's (\cite{achiam2023gpt}) powerful instruction-following capability to generate CoT rationales for each demonstration. These rationales are concatenated with the corresponding example images and provided as input to guide the model in using its rich knowledge for reasoning and simultaneously providing interpretable results.

Our primary contributions are summarized below.
(1) This paper synthetically explores the potential and promise of various paradigms of LVLMs applied to the field of traditional visual emotion recognition.
(2) We design a training-free framework to effectively apply the ICL paradigm to the CAER task. The proposed demonstration retrieval module retrieves demonstrations with the highest semantic similarity to the current test image from both the person context and scene context. This approach fully harnesses the ICL capability of LVLMs.
(3)
We conduct extensive experiments and perform a comprehensive analysis of the results to elucidate the impact of different paradigms on the emotion analysis performance of LVLMs. Our findings show that large models, even without parameter updates, can achieve competitive results in the CAER task, even surpassing traditional methods.


\section{Related Work}
\subsection{Context-aware Emotion Recognition}
The expression of emotions is often multifaceted. In visual emotion recognition tasks, facial expressions are commonly regarded as the most expressive visual information for conveying emotions, and much of the existing work has focused on facial expression analysis (\cite{jiang2020dfew}). However, in uncontrolled natural scenes, human emotions typically need to be inferred from a combination of visual information, including facial expressions, body movements, interactions, and the surrounding context. Recently, there have been numerous attempts to address context-aware emotion recognition tasks.~\cite{kosti2019context} introduced the EMOTIC dataset to encourage and support the Context-Aware Emotion Recognition (CAER) task, proposing a dual-stream convolutional architecture to separately extract information from human bodies and the entire scene. Similarly,~\cite{lee2019context} proposed a dual-stream architecture, but in contrast to EMOTIC, one branch is used to extract facial features while the other branch processes the entire image with the facial information masked as context.~\cite{yang2022emotion} released the HECO dataset for the CAER task and presented a novel multi-stream emotion recognition framework that incorporates four context information.~\cite{bose2023contextually} utilizes pre-trained vision-language models (VLMs) to extract descriptions of foreground context from images and proposes a multimodal context fusion module that combines foreground cues with visual scenes and human-centered contextual information for emotion prediction.~\cite{mittal2020emoticon} considered multiple signals such as facial expressions, postures, backgrounds, and depth to predict specific human emotions.~\cite{yang2023context,yang2024robust} introduced the causal modeling patterns to address spurious correlations between context and emotions caused by harmful context biases.

Although the above works have shown good performance on the CAER task, their capabilities are ultimately limited to the training datasets, resulting in poor generalization in real-world scenarios. Unlike these traditional training-based methods, we aim to explore the performance of various paradigms of LVLMs in the CAER task, seeking better approaches to overcome the limitations of traditional methods.

\subsection{Large Models in Affective Analysis}
As the scale of pre-trained language models continues to expand, the comprehension and generation capabilities of LLMs are rapidly advancing at an astonishing pace. Recently, several works have focused on exploring the emotion analysis capabilities of these large models.~\cite{lei2023instructerc} designed a retrieval template module and two emotional alignment tasks, utilizing fine-tuning to reform emotion recognition in conversation (ERC). Similarly,~\cite{zhang2023dialoguellm} had contributed to the development of large models in the ERC task. They fine-tuned the models using conversational context and emotional knowledge and incorporated multimodal information by converting videos into text for the fine-tuning process.~\cite{liu2024emollms} proposed a series of annotation tools for comprehensive affective analysis based on fine-tuning various LLMs with instruction data.

The aforementioned works are based on LLMs and primarily focus on emotion analysis in the field of NLP.~\cite{zhao2023prompting} proposed DFER-CLIP, which is based on the CLIP (\cite{radford2021learning}) model and specifically designed for dynamic facial expression recognition (DFER) in the wild.~\cite{cheng2024emotion} proposed Emotion-LLaMA, which integrates audio, visual, and text inputs for multimodal emotional recognition and reasoning.~\cite{lian2023gpt} comprehensively explored the performance of GPT-4 Vison (\cite{achiam2023gpt}) on generalized emotion recognition tasks.~\cite{xenos2024vllms} utilized LLaVA (\cite{liu2024visual}) to generate textual descriptions of images as auxiliary information for training a multimodal architecture for the CAER task, but it did not fully explore the emotion recognition capabilities of LVLMs. Different with~\cite{xenos2024vllms}, we are not using LVLMs as auxiliary tools, but rather exploring their inherent emotion analysis capabilities in the CAER task.

\section{Methodology}
\subsection{Proposed Framework for Few-Shot Setting}
\begin{figure}[htp]
\begin{center}
\includegraphics[width=1\textwidth]{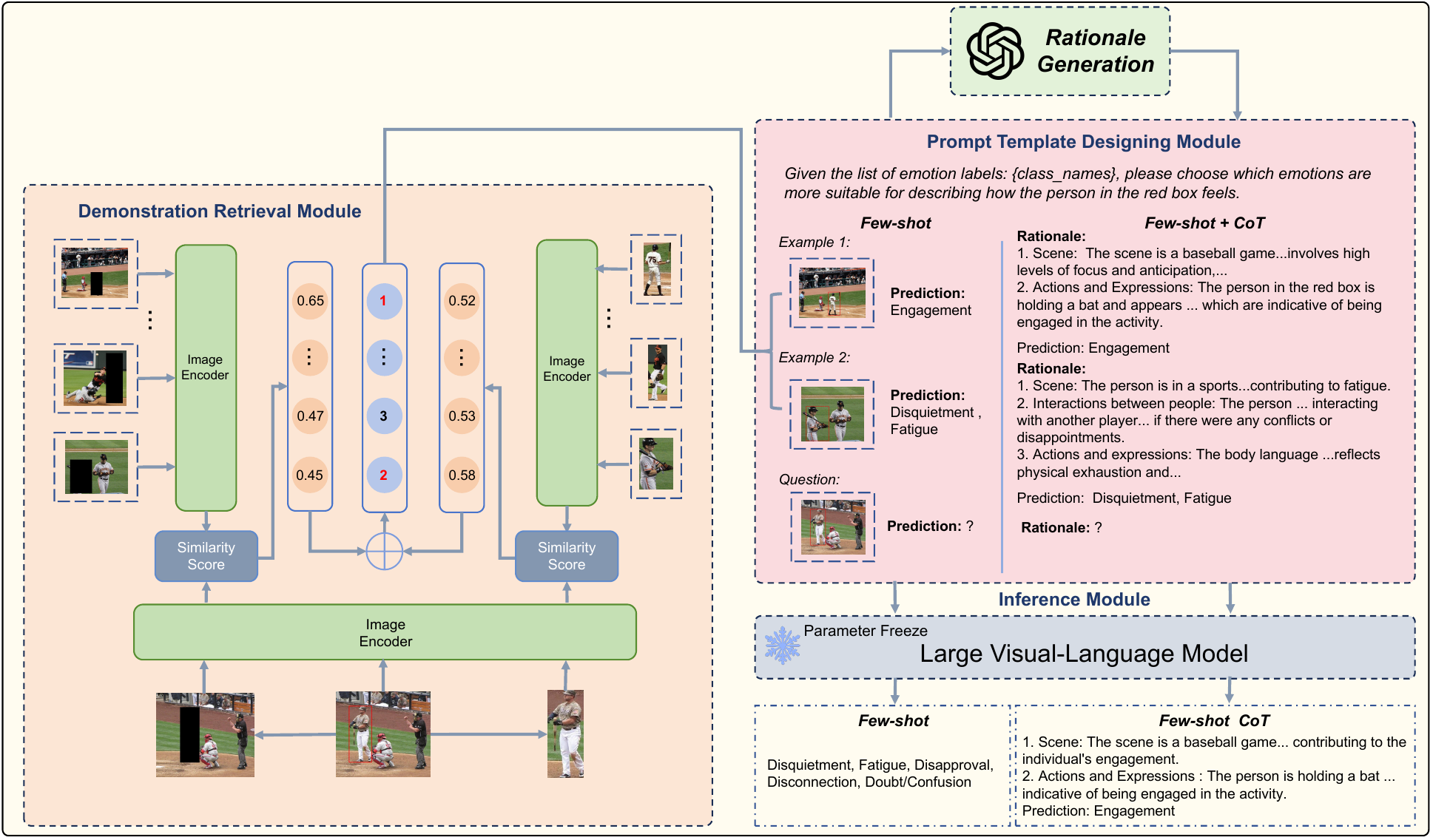}
\caption{The architecture of our framework on the 2-shot setting. Although we present the few-shot and few-shot CoT paradigms within the same framework, their inference processes are conducted separately in practice.}\label{fig:main_pic}
\end{center}
\vspace{-6pt}
\end{figure}

Figure~\ref{fig:main_pic} illustrates the overall architecture of our framework for the CAER task on the 2-shot setting. It is mainly composed of three parts: the \textit{Demonstration Retrieval Module}, the \textit{Prompt Template Designing Module}, and the \textit{Inference Module}.
\subsubsection{Demonstration Retrieval Module}
Previous works (\cite{zhang2024makes}) have emphasized the importance of demonstration selection in ICL and shown that examples with semantic similarity to the test sample can effectively enhance the performance of LVLMs. Traditional visual demonstration retrieval methods typically rely on image similarity. However, for the CAER task, considering only the overall image similarity is insufficient. On one hand, a single image may contain multiple annotated individuals, and each of them may express different emotions. Consequently, the images retrieved for different individuals should be distinct. On the other hand, the similarity between person context and scene context should be considered separately to prevent either context from being overlooked. For instance, when the proportion of the person context in an image is small, scene context may dominate the consideration, leading to the loss of important information. Therefore, we propose a demonstration retrieval module that jointly considers the similarity of person context and visual scene context. This approach ensures that both contexts are equally valued, facilitating the selection of the most appropriate demonstrations for different individuals within the same image.

\textbf{Person Context.} Facial expressions, gestures, body postures, and gait are all crucial indicators for assessing human emotions. We collectively refer to these indicators as the person context. Based on the provided bounding box annotations $[x_{min}, y_{min}, x_{max}, y_{max}]$, we crop the person from the image to obtain the person-specific context. Then, we use the image encoder (\textit{i.e.,} the pretrained ViT (\cite{dosovitskiy2020image})) to encode the cropped person and obtain the person-specific representations $f_{person}$,

\begin{equation}
f_{person} = ViT\left( I_{person} \right).
\end{equation}

\textbf{Scene Context.}
In addition to the explicit context of the individual, the surrounding scene semantics are essential for understanding emotions. To account for the influence of scene context, we designate the remaining part of the image $I$ after cropping out the labeled individual as the visual scene context $I_{scene}$,
\begin{equation}
f_{scene} = ViT\left( I_{scene} \right).
\end{equation}

\textbf{Image Similarity Rank.}
Given the test example $I_{q}$ and a set of candidate demonstration images $D=\{I_{1},I_{2},...,I_{N}\}$, where $I_{i}$ denotes the $i$-th candidate image. First, we crop each test image and candidate image into person and scene components, then encode these components to obtain their representations $F_{q}~=\left( f_{person\_ q},f_{scene\_ q} \right)$ and $F_{D} = \left\{ \left( f_{person\_ i},~f_{scene\_ i} \right) \right\}_{i=1} ^{N}$. Then we 
separately calculate the similarity of $F_{q}$ and $F_{D}$,

\begin{equation}
    {Sim}_{person_{i}} = ~\frac{f_{person\_ q}~ \cdot f_{person\_ i}^{T}}{\left\| f_{person\_ q} \right\|\left\| f_{person\_ i} \right\|},
\end{equation}

\begin{equation}
    {Sim}_{scene_{i}} = ~\frac{f_{scene\_ q}~ \cdot f_{scene\_ i}^{T}}{\left\| f_{scene\_ q} \right\|\left\| f_{scene\_ i} \right\|},
\end{equation}

\begin{equation}
    Sim\left( F_{q},F_{D} \right) = \left\{ \left({Sim}_{person_{i}}+ {Sim}_{scene_{i}}~ \right)/2 \right\}_{i=1} ^{N},
\end{equation}
and then record the rank $R_{q~}\in \mathbb{R}^N $ of each candidate image in $D$,
\begin{equation}
    R_{q~} = ~Rank\left( Sim\left( F_{q},F_{D} \right) \right).
\end{equation}
Compared to ranking based on overall image similarity scores, the method of ranking based on the similarity scores of two distinct contexts better accounts for the influence of different contexts on the emotional semantics of images, thereby providing optimal demonstrations for the ICL learning of LVLMs. Finally, we take the top-k images with the highest similarity ranking as selected demonstrations.

\subsubsection{Prompt Template Designing Module}
\label{sec3.1.2}
To better transfer the capabilities of LVLMs to the CAER task, we reframe the CAER task as a generative task. We construct a prompt template to bridge the gap when applying LVLMs to a subtask. As shown in Figure~\ref{fig:main_pic}, for the CAER task, each input consists of three parts: Instruction, Demonstration, and Test Sample.

\textbf{Instruction.}
At the beginning of the input, we first provide LVLM with a clear task instruction and a label statement as follows: ``Given the list of emotion labels: \{class names\}, please choose which emotion is more suitable for describing how the person in the red box feels'', where \{class names\} are substituted for the list of class names available in each dataset. Specifically, for multi-label classification tasks, we replace `emotion is' with `emotions are' to prevent LVLMs from producing a single-label output.

\begin{figure}[ht]
\begin{center}
\includegraphics[width=1\textwidth]{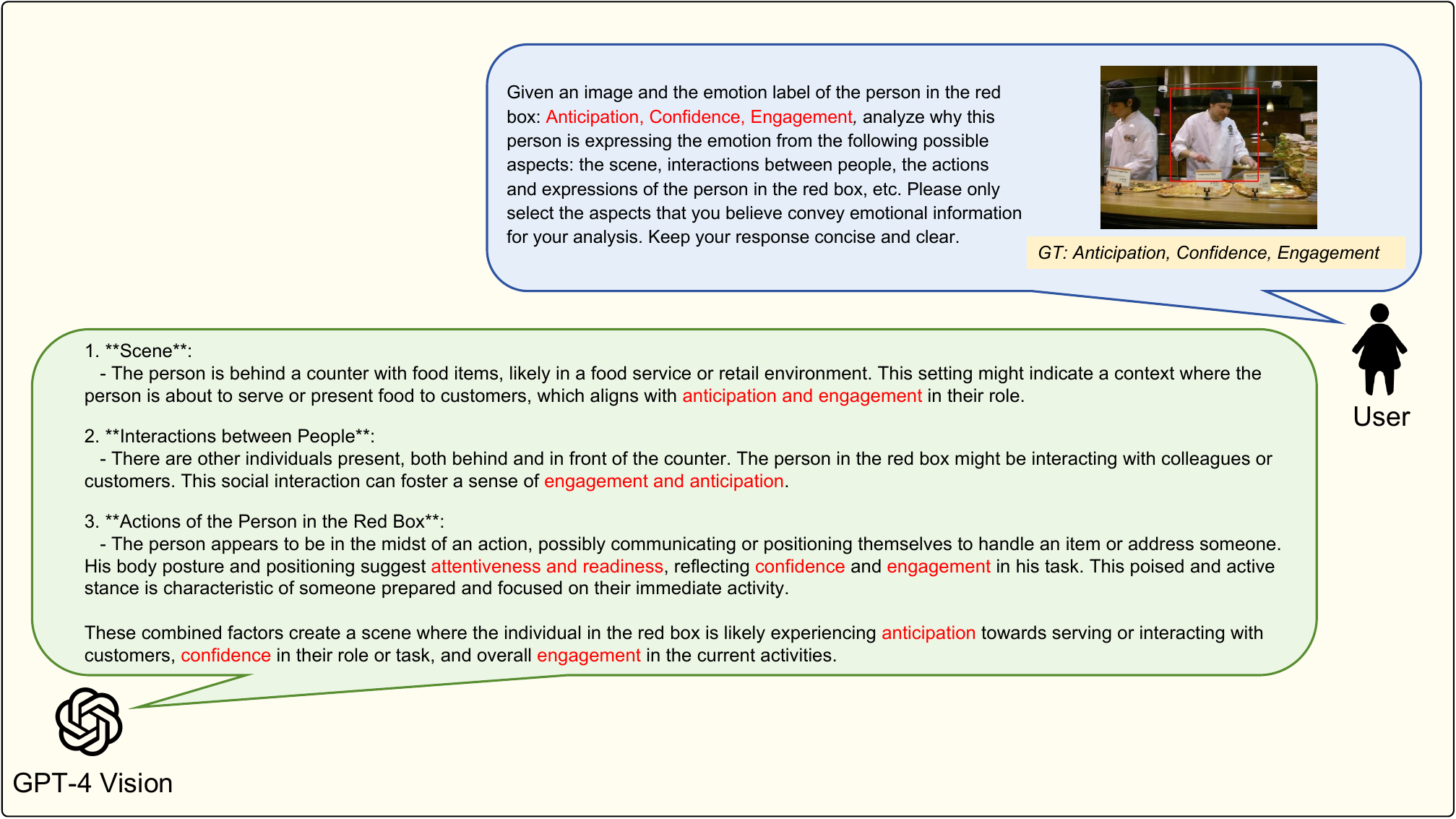}
\caption{An example of emotional rationale generated for the EMOTIC dataset using GPT-4 Vision. The emotion-related words are highlighted in red.}\label{fig:GPT4_example}
\end{center}
\vspace{-10pt}
\end{figure}

\textbf{Demonstration.}
Each of our demonstrations is an image-answering pair $\{I, A\}$. We select the top-k demonstrations obtained by the proposed demonstration retrieval module and concatenate them to form the overall demonstration.
\begin{equation}
    ~\sigma = ~Top\_ K\left( R_{q~} \right),
\end{equation}
\begin{equation}
    {demon}_{i}~ = ~\left\{ {F_{D}}^{\sigma_{i}} ,y_{\sigma_{i}} \right\},
\end{equation}
where $\sigma$ are the indices of top-k similarity ranking, and $\sigma = \{ \sigma_{1},...,\sigma_{k}\}$. Specifically, for the few-shot CoT paradigm, our demonstrations include not only the images and labels but also the rationales that explain why the image corresponds to the given emotion labels,
\begin{equation}
    {demon}_{i}~ = ~\left\{ {F_{D}}^{\sigma_{i}},{Rationale\left( {F_{D}}^{\sigma_{i}},y_{\sigma_{i}} \right), y}_{\sigma_{i}} \right\}.
\end{equation}
The rationale is generated by GPT-4 Vision (\cite{achiam2023gpt}) due to its powerful instruction-following capabilities. Figure~\ref{fig:GPT4_example} shows an example of generated rationale.
The final demonstration is expressed as follows:
\begin{equation}
    Demonstration~ = ~Concat\left({demon}_{1},...,{demon}_{k}\right).
\end{equation}

\subsubsection{Inference Module}
Following the instruction and demonstration, we append the test sample at the end of the prompt as input to the LVLM and select the most likely generated sequence as the output,
\begin{equation}
    {Input}_{q}~ = ~\left\{ Instruction, Demonstration, I_{q} \right\},
\end{equation}
\begin{equation}
    {Output}_{q~} = LVLM({Input}_{q},\theta).
\end{equation}
Finally, we post-process the output sequence to obtain the list of predicted labels.

\section{Experiments}
\subsection{Dataset and Evaluation Metrics}
EMOTIC (\cite{kosti2019context}) is the most widely used context emotion dataset that collected samples in uncontrolled environments. It contains 23,571 images of 34,320 annotated agents, and each individual is assigned one or more of the 26 discrete emotion labels based on their apparent emotional states. The dataset is randomly split into training (70\%), validation (10\%), and testing (20\%) sets. The training set is annotated by a single annotator, whereas the validation and testing sets are annotated by four and two additional annotators, respectively. Consequently, the average number of labels per sample is 1.96 for the training set, 6.15 for the validation set, and 4.42 for the testing set.

HECO (\cite{yang2022emotion}) dataset consists of images from the HOI (\cite{chao2018learning}) datasets, film clips, and images from the Internet. The dataset contains 9,385 images of 19,781 annotated agents. Each agent is labeled with one of the eight emotions: Surprise, Excitement, Happiness, Peace, Disgust, Anger, Fear, and Sadness. The dataset is randomly split into training (70\%), validation (10\%), and testing (20\%) sets.

Since the LVLM treats the CAER task as a text generation process, its output consists of textual emotion labels rather than probabilities. Consequently, we cannot use mAP (mean Average Precision) as an evaluation metric, as was done in previous works~(\cite{kosti2019context,mittal2020emoticon}). In this work, we report Precision, Recall, F1 Score, Hamming Loss, and multi-label Accuracy on the EMOTIC (both micro average and macro average). For the HECO, we use the Precision, Recall, F1 Score on macro average, and the standard classification accuracy for evaluation. Note that a smaller Hamming Loss and larger Accuracy, Precision, Recall, and F1 Score indicate better classification quality.

\subsection{Experimental Setup}
Extensive experiments are conducted on two LVLMs, including LLAVA-7B (\cite{liu2024visual}) and VILA-8B (\cite{lin2024vila}). We find that LLAVA struggles with handling multiple image inputs, and its instruction-following capability significantly decreases as the number of input images increases. Therefore, for the few-shot setting, we only report the 2-shot results for LLAVA. For the EMOTIC and HECO datasets, We randomly select 200 examples from the training set as the candidate demonstrations set and chose $k$ examples from the set for the $k$-shot ICL task. Specifically, due to the difference in the average number of labels between the EMOTIC training set and validation set, we additionally randomly sample 200 examples from the EMOTIC validation set as a candidate set to assess the impact of the number of demonstration labels on few-shot prediction results.
To facilitate subsequent processing of the generated text, we specify ``reply with label(s) only'' for zero-shot prompts, while for few-shot tasks, the output format is demonstrated in the examples.
All experiments are conducted on four Nvidia A800-80G GPUs. For the fine-tuning approach, we set the learning rate to 1e-5 and the batch size to 8.

\renewcommand{\arraystretch}{1.5}
\begin{table}[ht]
\caption{
Comparison results on the EMOTIC dataset. 
The left and right of ``$/$'' denote the micro and macro average results, respectively.
``SFT'' is the supervised fine-tuning. For the few-shot CoT setting, we report the best results among different numbers of demonstrations.}
\label{main_table_emotic}
\resizebox{\textwidth}{!}{
\begin{tabular}{lcccccc}
\toprule

\textbf{Setting}                            & \textbf{Models}    & \textbf{Precision} $\uparrow$ & \textbf{Recall} $\uparrow$ & \textbf{F1 Score} $\uparrow$ & \textbf{Hamming} $\downarrow$ & \textbf{Accuracy} $\uparrow$ \\ \hline

Trained                            & EMOT-Net   &     46.00/21.40      &   45.96/21.38     &     45.98/21.39      &     18.34        &      \textbf{2.90}    \\ \hline
\multirow{2}{*}{Zero-shot}         & LLaVA-7B &     34.55/24.50      &   11.65/11.00     &    17.43/11.19       &    18.75         &    0.19   \\
                                   & VILA-8B  &    24.32/30.27       &    42.17/43.43    &     30.84/21.49      &      32.10      &      0.40    \\ \hline
\multirow{2}{*}{SFT}               & LLaVA-7B    &  \textbf{80.20/57.81}     &   23.81/11.04     &    36.72/15.46      &       13.94       &    1.00      \\
                                   & VILA-8B     &     76.85/53.02      &   27.53/15.51     &    41.94/19.76       &       \textbf{13.56}       &    1.10        \\\hline
\multirow{3}{*}{Few-shot}          & LLaVA-7B  &     34.94/22.86      &    26.27/20.10    &    29.99/18.51      &       20.83       &    0.86     \\ 
                                   & VILA-8B (Train Subset, 10-shot)  &    43.21/32.36       &   43.01/31.33     &    43.11/26.70       &       19.28       &      1.10    \\ 
                                   & VILA-8B (Val Subset, 10-shot)  &    37.88/27.86       &   \textbf{73.31/46.90}     &    \textbf{49.95/29.81}       &       24.94       &      0.25    \\ \hline
\multirow{3}{*}{Few-shot CoT}      & LLaVA-7B  &  48.49/26.34         &    13.17/8.81    &       20.71/12.21    &       17.12       &      0.28    \\ 
                                   & VILA-8B (Train Subset, 8-shot)  &    57.82/40.73       &   24.98/16.13     &    34.88/19.87       &       15.83       &      0.82    \\ 
                                   & VILA-8B (Val Subset, 4-shot)  &     41.24/23.96      &   56.16/34.04     &     47.56/26.69      &       21.03       &     0.44     \\ \bottomrule

\end{tabular}
}
\end{table}

\subsection{Results and Analysis}

The results of zero-shot setting, supervised fine-tuning (SFT) setting, few-shot setting, and few-shot CoT setting are shown in Table~\ref{main_table_emotic} and Table~\ref{main_table_heco}. Due to the differing task characteristics and complexities of the EMOTIC and HECO (\textit{i.e.,} EMOTIC being a multi-label classification task and HECO a single-label classification task), we analyze the results of the two datasets separately.

For the EMOTIC, the fine-tuned LLAVA achieves the highest precision. VILA performs few-shot learning by retrieving demonstrations from the validation subset, obtaining the highest recall and F1 score. Compared to the trained EMOT-Net (\cite{kosti2017emotion}), the fine-tuned LVLMs exhibit significantly higher precision but considerably lower recall. This suggests that the fine-tuned LVLMs tend to output the most accurate emotions rather than the most comprehensive ones because the average number of labels in the EMOTIC training set is much smaller than that in the test set. Although EMOT-Net is also trained on the training set, it outputs probability values, unlike LVLMs which generate direct labels. This allows for dynamic threshold adjustment to balance precision and recall, preventing the imbalance seen in LVLMs.

Another notable observation is the significant gap between the micro average and macro average scores for both the trained EMOT-Net and the fine-tuned LVLMs. The lower macro average scores indicate poor performance on certain emotion categories, while the higher micro average scores suggest that the well-performing emotion categories are more prevalent in the dataset. This discrepancy highlights the issue of imbalanced label distribution in the dataset, consistent with previous findings (\cite{yang2024robust}). Whether through training or fine-tuning, models are profoundly influenced by the distribution of the training set. For subjective annotations like emotions, this is not ideal. Focusing on fitting the training set's label distribution causes models to learn the annotators' emotional biases, and the imbalanced label distribution leads to insufficient learning of less prevalent emotions.

Furthermore, we observe the performance of LVLMs in the few-shot CoT paradigm. We find that the addition of rationale examples increases precision but decreases recall. That's because without CoT, the model likely recognizes simple patterns, but with CoT examples, it needs to learn and focus on more detailed information. This makes the model more stringent and cautious in emotion classification, leading to the omission of some true labels. How to streamline the CoT process, allowing LVLMs to balance resources between analysis and decision-making during inference, will be the focus of our future research.

\begin{table}[h]
\caption{Comparison results on the HECO dataset.}
\centering
\label{main_table_heco}
\resizebox{0.8\textwidth}{!}{
\begin{tabular}{lccccc}
\toprule
\textbf{Setting}                            & \textbf{Models}     & \textbf{Precision} $\uparrow$ & \textbf{Recall} $\uparrow$ & \textbf{F1 Score} $\uparrow$ & \textbf{Accuracy} $\uparrow$ \\ \hline
Trained                            & EMOT-Net    &  10.20         &    13.88    &     9.89      &     38.80      \\ \hline
\multirow{2}{*}{Zero-shot}         & LLaVA-7B  &     32.57      &   25.87     &     17.71      &      35.69     \\
                                   & VILA-8B   &     42.54      &    25.72    &      21.27     &     35.56     \\ \hline
\multirow{2}{*}{SFT}               & LLaVA-7B     &   46.98        &    28.47    &    29.90       &     57.16      \\
                                   & VILA-8B      &    \textbf{48.31 }      &    \textbf{33.62}    &     \textbf{35.81}  &     \textbf{60.82}   \\\hline
\multirow{2}{*}{Few-shot}          & LLaVA-7B (2-shot)   &   19.30        &   19.64     &    15.50       &    32.27       \\ 
                                   & VILA-8B (10-shot)   &     37.26      &    28.15    &    27.57       &    47.55       \\ \hline
\multirow{2}{*}{Few-shot CoT}      & LLaVA-7B (2-shot)   &    9.05      &      5.23  &       4.94    &    20.13       \\ 
                                   & VILA-8B (8-shot)   &     25.19      &   22.38     &    19.53       &    39.80       \\ \bottomrule
\end{tabular}
}
\end{table}

In the zero-shot setting, we find that unguided or non-finetuned LVLMs perform poorly on the CAER task. However, we also notice that the gap between the micro average score and the macro average score is not significant, indicating that untrained LVLMs do not exhibit a clear bias toward specific emotions. 
In the few-shot paradigm, the demonstrations provided to LVLMs significantly enhance their emotion recognition capabilities. VILA achieves performance comparable to EMOT-Net in the few-shot setting where the retrieval set is derived from the training set. It also shows a better balance between precision and recall, with the gap between micro and macro average scores narrowing considerably. Due to LLAVA's relatively poor ICL capability, its performance in the few-shot setting is subpar. This suggests that our work will be limited by the inherent capabilities of the LVLMs. However, the good results of VILA indicate that LVLMs with good ICL ability can achieve relatively good and balanced emotion analysis performance with just few-shot prompts. When the retrieval set is derived from the validation set, VILA's performance in the F1 Score significantly surpasses the EMOT-Net. We believe this is because the average number of labels in the validation set is closer to that of the test set, resulting in a much higher recall. This also shows that untuned LVLMs still rely on the quality of samples in the retrieval set. However, compared to the training dataset, which contains tens of thousands of samples, the retrieval set is much smaller. Maintaining a high-quality retrieval set is far less costly than maintaining a good training set. More importantly, the few-shot paradigm requires no training, meaning the cost of transferring LVLMs to different scenarios is minimal. This demonstrates the potential of LVLMs in the CAER task, paving the way for promising future developments in this field.

For the HECO, since it is a single-label classification task with only 8 emotion categories (much fewer than 26 categories from EMOTIC), the task is simpler, and the issues of emotional bias observed with the EMOTIC are less likely to be reflected in HECO. As shown in Table~\ref{main_table_heco}, the supervised fine-tuned LVLMs outperform traditional models and all other paradigms. Even in the zero-shot setting, LVLMs show performance comparable to EMOT-Net. In the few-shot paradigm, VILA achieves a significantly higher accuracy of 47.55\% compared to 38.80\% of EMOT-Net. However, we found that although the few-shot paradigm outperforms the zero-shot paradigm and traditional models, it did not surpass the SFT paradigm, which is contrary to our findings in the EMOTIC dataset. This is because the label distribution in the training set is consistent with that in the test set in HECO, allowing the model to generalize well from the training set to the testing set through SFT. Furthermore, when we add CoT, the model performance decreases compared to not using CoT. The CoT method is more suitable for complex, multi-step reasoning tasks. In single-label emotion classification, the decision process is usually straightforward and does not require complex reasoning steps. The CoT method introduces additional reasoning steps and explanations, which may over-complicate the task, adding unnecessary computation and decision-making steps, thus reducing the performance weakening.

\subsection{Ablation Studies}       

\begin{figure}[t]
\begin{center}
\includegraphics[width=0.9\textwidth]{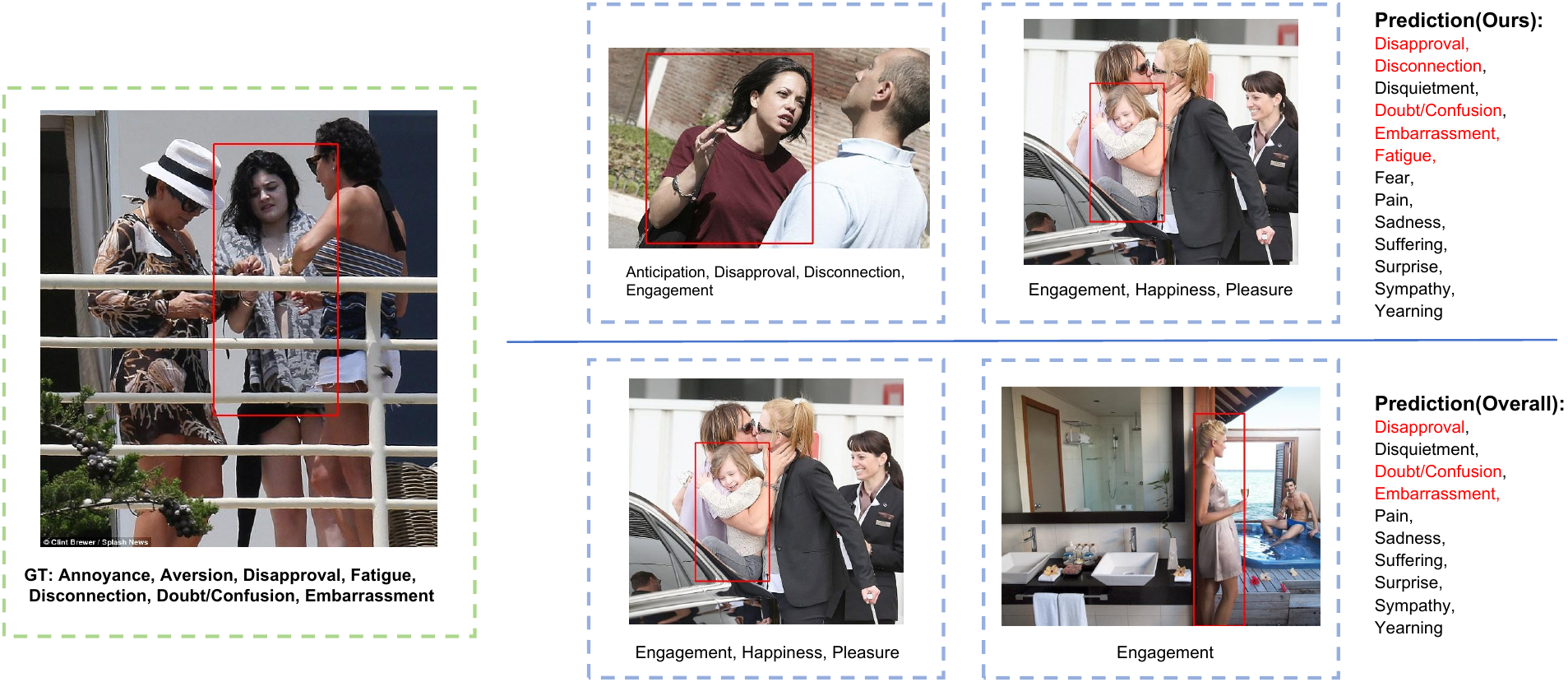}
\caption{An example of the results obtained using different retrieval methods. The labels under the images represent the ground truth (GT) for each image. The image on the left is the test sample, while the upper right shows the demonstrations retrieved by our method, and the lower right shows the demonstrations retrieved based on overall image similarity.}\label{fig:retrieval}
\end{center}
\vspace{-10pt}
\end{figure}

\textbf{Impact of Demonstration Retrieval Method.}
We report the results of five retrieval ranking methods under the 6-shot setting in Table~\ref{retrieve method ablation}. In this case, ``Overall'' refers to ranking based on the similarity of the entire image, ``body-based'' refers to calculating similarity using only the cropped body part, ``scene-based'' refers to calculating similarity using only the scene part, and ``random'' involves randomly selecting images as demonstrations. 
We observe that the proposed multi-context retrieval method performs the best, followed by overall image retrieval, while the random selection method performs the worst. This indicates that separately considering person context and scene context is beneficial for the CAER task. Furthermore, considering the similarity of the entire image is more effective than only considering the person context or scene context, demonstrating that each context contributes to retrieving the optimal demonstrations.

\begin{table}[h]
\caption{Comparison results of different retrieval methods on EMOTIC and HECO. }
\centering
\label{retrieve method ablation}
\resizebox{0.7\textwidth}{!}{
\begin{tabular}{lcccc}
\toprule
\multirow{2}{*}{\textbf{Methods}}    & \multicolumn{3}{c}{\textbf{EMOTIC}}                         & \multicolumn{1}{c}{\textbf{HECO}} \\ \cline{2-5}
            & \textbf{F1 Score} $\uparrow$ & \textbf{Hamming} $\downarrow$ & \textbf{Accuracy} $\uparrow$   & \textbf{Accuracy} $\uparrow$\\ \hline
Random      &  35.72/23.20  &    24.45     & 0.35    &    43.96    \\ 
Scene-base  &  38.00/24.22  &    22.39     & 0.74    &    42.99    \\ 
Body-base   &  39.52/25.04  &    22.10     & 0.89    &    44.46    \\ 
Overall     &  40.30/25.52  &    21.42     & 0.87    &    43.58    \\ 
Ours        &  \textbf{40.41/25.93}  &   \textbf{21.37}      &   \textbf{0.92}  &    \textbf{46.74}    \\ \bottomrule
\end{tabular}
}
\end{table}

Figure~\ref{fig:retrieval} presents the top 2 demonstrations obtained by our retrieval method and the ``Overall'' retrieval method, along with the prediction results based on these demonstrations. We can see that images retrieved by our method are semantically closer to the test samples (\textit{i.e.,} both expressing negative emotions), whereas the images retrieved based on overall image similarity all express positive emotions. The prediction results indicate that the samples retrieved by our method help the model predict more correct labels.


\begin{table}[t]
\caption{Comparison results of different shot settings on EMOTIC and HECO. }
\centering
\label{n-shot ablation}
\resizebox{0.7\textwidth}{!}{
\begin{tabular}{lcccc}
\toprule
\multirow{2}{*}{\textbf{Setting}}   & \multicolumn{3}{c}{\textbf{EMOTIC}}                         & \multicolumn{1}{c}{\textbf{HECO}} \\ \cline{2-5}
                           &  \textbf{F1 Score}  $\uparrow$ & \textbf{Hamming} $\downarrow$ & \textbf{Accuracy} $\uparrow$ & \textbf{Accuracy} $\uparrow$\\ \hline
0-shot   &      30.84/21.49      &      32.10      &      0.40  &   35.56    \\ 
2-shot      &     37.88/25.33      &       24.18       &   0.79    &  40.09  \\                       
4-shot      &   39.30/25.76         &        23.31      &     0.69  &  42.07  \\ 
6-shot      &   40.41/25.93        &     21.57         &    0.92   &  46.74  \\                         
8-shot      &      41.71/26.70      &         19.84     &     0.99  &  47.22  \\                      
10-shot     &      43.11/\textbf{26.70}      &       \textbf{19.28}  &     1.10  &  \textbf{47.55}  \\ 
12-shot     &     43.19/26.25      &       19.52       &     1.25 &   44.93  \\ 
16-shot     &      \textbf{43.71}/25.98      &       19.38       &    \textbf{1.44}  &  41.47  \\ \bottomrule
                         
\end{tabular}
}
\end{table}

\textbf{Impact of Demonstration Number.}
To investigate the impact of the number of demonstrations on performance, we conduct experiments with varying numbers of examples, as shown in Table~\ref{n-shot ablation}. We observe that, on both datasets, most metrics generally improve as the number of examples increases. Our framework performs best on average with 10-shot demonstrations. Beyond 10-shot, the improvements stabilize and some metrics even decline. This decline can be attributed to the fact that longer inputs make it difficult for the model to focus on key points, and more demonstrations may include semantically dissimilar examples, which can have a negative effect. Therefore, selecting an appropriate number of demonstrations is crucial for maximizing the potential of LVLMs.

\begin{figure}[t]
\begin{center}
\includegraphics[width=1\textwidth]{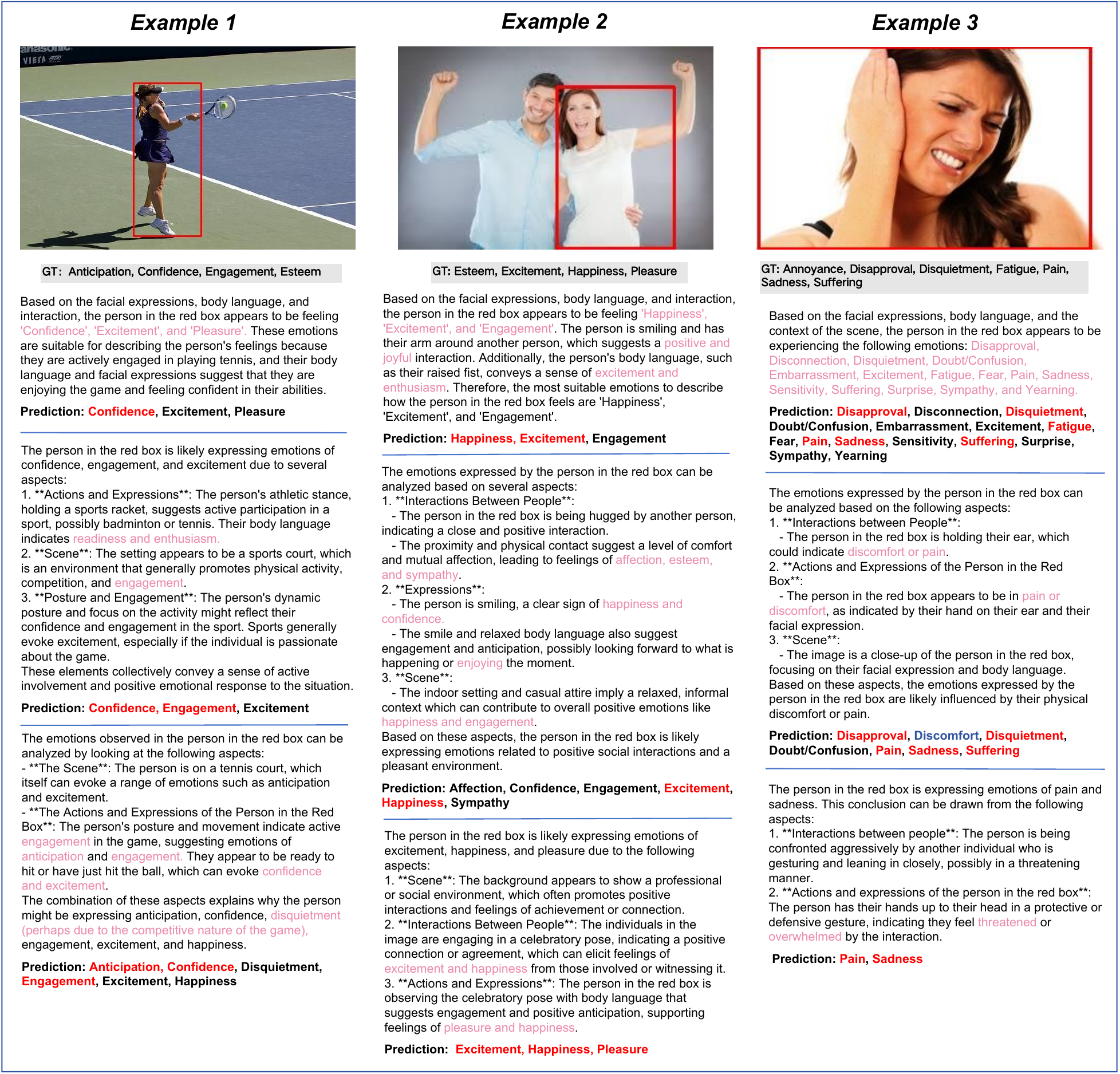}
\caption{Examples of the CoT results on the EMOTIC. Each example from top to bottom shows generations in 0-shot CoT, 2-shot CoT (demonstrations from the validation subset), and 2-shot CoT (demonstrations from the training subset) settings. The pink text highlights the emotions mentioned in the rationale, the red labels indicate correctly predicted labels, and the blue text denotes labels that are not within the GT range.}\label{fig:case_study}
\end{center}
\vspace{-5pt}
\end{figure}

\subsection{Case Study}
Figure~\ref{fig:case_study} presents examples of the CoT results under different settings on the EMOTIC. We find that in the zero-shot setting, while the model's decisions are not poor, the generated rationale does not explain in detail why the person feels this emotion, and in some cases (\textit{e.g.,} Example 3), it is merely a redundant description of the decision. With few demonstrations, the model's CoT improves significantly by learning from the demonstrated CoT to generate rationale from different perspectives. And it often leads to more accurate decisions. CoT with images from the validation subset tends to predict more labels, whereas CoT with images from the training set yields more concise results. We can achieve our desired outcomes by constructing retrieval sets with different characteristics for the model.

Additionally, we find that although sometimes the model predictions are not always entirely correct, its analysis is subjectively reasonable. For instance, in the first image, the model thinks the person should feel ``Excitement'' in a sports setting, which seems understandable even though this category is not among the ground truth labels of the image. In contrast, the ``Esteem'' label in the ground truth is not clearly perceivable, which we believe reflects the annotator's emotional bias. For the third image, the model provided the label "Discomfort", which is not one of the 26 emotion categories specified in the EMOTIC dataset. However, emotions are inherently complex and cannot be fully encapsulated by a limited set of categories. Models trained on specific datasets might ignore or consider ``Discomfort'' as an incorrect decision, but as humans, we recognize it as a right label. This underscores the necessity of our research: emotion analysis should be comprehensive rather than limited. Traditional methods cannot achieve this, whereas large models hold great potential in this regard.

\section{Conclusion}
This paper explores the potential and performance of various LVLM paradigms in the CAER task. Our findings indicate that LVLMs can outperform traditional models without additional training through our proposed framework, demonstrating their significant potential in understanding contextual emotions. The powerful reasoning and generalization capabilities of large models provide a promising outlook: we envision that emotion analysis should not be constrained by the limitations of specific datasets but should advance toward a broader and more fine-grained emotional landscape. 

\noindent \textbf{Future Work.} We plan to improve the performance of LVLMs in the Few-shot CoT paradigm by constructing higher-quality CoT demonstrations, guiding the model to generate more logical and comprehensive rationales.

\acks{This work is supported by the National Key R\&D Program of China under Grant 2021ZD0113502.}

\bibliography{acml24}

\begin{thebibliography}{29}
\providecommand{\natexlab}[1]{#1}
\providecommand{\url}[1]{\texttt{#1}}
\expandafter\ifx\csname urlstyle\endcsname\relax
  \providecommand{\doi}[1]{doi: #1}\else
  \providecommand{\doi}{doi: \begingroup \urlstyle{rm}\Url}\fi

\bibitem[Achiam et~al.(2023)Achiam, Adler, Agarwal, Ahmad, Akkaya, Aleman, Almeida, Altenschmidt, Altman, Anadkat, et~al.]{achiam2023gpt}
Josh Achiam, Steven Adler, Sandhini Agarwal, Lama Ahmad, Ilge Akkaya, Florencia~Leoni Aleman, Diogo Almeida, Janko Altenschmidt, Sam Altman, Shyamal Anadkat, et~al.
\newblock Gpt-4 technical report.
\newblock \emph{arXiv preprint arXiv:2303.08774}, 2023.

\bibitem[Alayrac et~al.(2022)Alayrac, Donahue, Luc, Miech, Barr, Hasson, Lenc, Mensch, Millican, Reynolds, et~al.]{alayrac2022flamingo}
Jean-Baptiste Alayrac, Jeff Donahue, Pauline Luc, Antoine Miech, Iain Barr, Yana Hasson, Karel Lenc, Arthur Mensch, Katherine Millican, Malcolm Reynolds, et~al.
\newblock Flamingo: a visual language model for few-shot learning.
\newblock \emph{Advances in neural information processing systems}, 35:\penalty0 23716--23736, 2022.

\bibitem[Bose et~al.(2023)Bose, Hebbar, Somandepalli, and Narayanan]{bose2023contextually}
Digbalay Bose, Rajat Hebbar, Krishna Somandepalli, and Shrikanth Narayanan.
\newblock Contextually-rich human affect perception using multimodal scene information.
\newblock In \emph{ICASSP 2023-2023 IEEE International Conference on Acoustics, Speech and Signal Processing (ICASSP)}, pages 1--5. IEEE, 2023.

\bibitem[Brown et~al.(2020)Brown, Mann, Ryder, Subbiah, Kaplan, Dhariwal, Neelakantan, Shyam, Sastry, Askell, et~al.]{brown2020language}
Tom Brown, Benjamin Mann, Nick Ryder, Melanie Subbiah, Jared~D Kaplan, Prafulla Dhariwal, Arvind Neelakantan, Pranav Shyam, Girish Sastry, Amanda Askell, et~al.
\newblock Language models are few-shot learners.
\newblock \emph{Advances in neural information processing systems}, 33:\penalty0 1877--1901, 2020.

\bibitem[Chao et~al.(2018)Chao, Liu, Liu, Zeng, and Deng]{chao2018learning}
Yu-Wei Chao, Yunfan Liu, Xieyang Liu, Huayi Zeng, and Jia Deng.
\newblock Learning to detect human-object interactions.
\newblock In \emph{2018 ieee winter conference on applications of computer vision (wacv)}, pages 381--389. IEEE, 2018.

\bibitem[Cheng et~al.(2024)Cheng, Cheng, He, Sun, Wang, Lin, Lian, Peng, and Hauptmann]{cheng2024emotion}
Zebang Cheng, Zhi-Qi Cheng, Jun-Yan He, Jingdong Sun, Kai Wang, Yuxiang Lin, Zheng Lian, Xiaojiang Peng, and Alexander Hauptmann.
\newblock Emotion-llama: Multimodal emotion recognition and reasoning with instruction tuning.
\newblock \emph{arXiv preprint arXiv:2406.11161}, 2024.

\bibitem[Dosovitskiy et~al.(2020)Dosovitskiy, Beyer, Kolesnikov, Weissenborn, Zhai, Unterthiner, Dehghani, Minderer, Heigold, Gelly, et~al.]{dosovitskiy2020image}
Alexey Dosovitskiy, Lucas Beyer, Alexander Kolesnikov, Dirk Weissenborn, Xiaohua Zhai, Thomas Unterthiner, Mostafa Dehghani, Matthias Minderer, Georg Heigold, Sylvain Gelly, et~al.
\newblock An image is worth 16x16 words: Transformers for image recognition at scale.
\newblock \emph{arXiv preprint arXiv:2010.11929}, 2020.

\bibitem[Jiang et~al.(2020)Jiang, Zong, Zheng, Tang, Xia, Lu, and Liu]{jiang2020dfew}
Xingxun Jiang, Yuan Zong, Wenming Zheng, Chuangao Tang, Wanchuang Xia, Cheng Lu, and Jiateng Liu.
\newblock Dfew: A large-scale database for recognizing dynamic facial expressions in the wild.
\newblock In \emph{Proceedings of the 28th ACM international conference on multimedia}, pages 2881--2889, 2020.

\bibitem[Kosti et~al.(2017)Kosti, Alvarez, Recasens, and Lapedriza]{kosti2017emotion}
Ronak Kosti, Jose~M Alvarez, Adria Recasens, and Agata Lapedriza.
\newblock Emotion recognition in context.
\newblock In \emph{Proceedings of the IEEE conference on computer vision and pattern recognition}, pages 1667--1675, 2017.

\bibitem[Kosti et~al.(2019)Kosti, Alvarez, Recasens, and Lapedriza]{kosti2019context}
Ronak Kosti, Jose~M Alvarez, Adria Recasens, and Agata Lapedriza.
\newblock Context based emotion recognition using emotic dataset.
\newblock \emph{IEEE transactions on pattern analysis and machine intelligence}, 42\penalty0 (11):\penalty0 2755--2766, 2019.

\bibitem[Lee et~al.(2019)Lee, Kim, Kim, Park, and Sohn]{lee2019context}
Jiyoung Lee, Seungryong Kim, Sunok Kim, Jungin Park, and Kwanghoon Sohn.
\newblock Context-aware emotion recognition networks.
\newblock In \emph{Proceedings of the IEEE/CVF international conference on computer vision}, pages 10143--10152, 2019.

\bibitem[Lei et~al.(2023)Lei, Dong, Wang, Wang, and Wang]{lei2023instructerc}
Shanglin Lei, Guanting Dong, Xiaoping Wang, Keheng Wang, and Sirui Wang.
\newblock Instructerc: Reforming emotion recognition in conversation with a retrieval multi-task llms framework.
\newblock \emph{arXiv preprint arXiv:2309.11911}, 2023.

\bibitem[Lian et~al.(2023)Lian, Sun, Sun, Chen, Wen, Gu, Chen, Liu, and Tao]{lian2023gpt}
Zheng Lian, Licai Sun, Haiyang Sun, Kang Chen, Zhuofan Wen, Hao Gu, Shun Chen, Bin Liu, and Jianhua Tao.
\newblock Gpt-4v with emotion: a zero-shot benchmark for multimodal emotion understanding.
\newblock \emph{arXiv preprint arXiv:2312.04293}, 2023.

\bibitem[Lin et~al.(2024)Lin, Yin, Ping, Molchanov, Shoeybi, and Han]{lin2024vila}
Ji~Lin, Hongxu Yin, Wei Ping, Pavlo Molchanov, Mohammad Shoeybi, and Song Han.
\newblock Vila: On pre-training for visual language models.
\newblock In \emph{Proceedings of the IEEE/CVF Conference on Computer Vision and Pattern Recognition}, pages 26689--26699, 2024.

\bibitem[Liu et~al.(2024{\natexlab{a}})Liu, Li, Wu, and Lee]{liu2024visual}
Haotian Liu, Chunyuan Li, Qingyang Wu, and Yong~Jae Lee.
\newblock Visual instruction tuning.
\newblock \emph{Advances in neural information processing systems}, 36, 2024{\natexlab{a}}.

\bibitem[Liu et~al.(2024{\natexlab{b}})Liu, Yang, Zhang, Xie, Yu, and Ananiadou]{liu2024emollms}
Zhiwei Liu, Kailai Yang, Tianlin Zhang, Qianqian Xie, Zeping Yu, and Sophia Ananiadou.
\newblock Emollms: A series of emotional large language models and annotation tools for comprehensive affective analysis.
\newblock \emph{arXiv preprint arXiv:2401.08508}, 2024{\natexlab{b}}.

\bibitem[Mittal et~al.(2020)Mittal, Guhan, Bhattacharya, Chandra, Bera, and Manocha]{mittal2020emoticon}
Trisha Mittal, Pooja Guhan, Uttaran Bhattacharya, Rohan Chandra, Aniket Bera, and Dinesh Manocha.
\newblock Emoticon: Context-aware multimodal emotion recognition using frege's principle.
\newblock In \emph{Proceedings of the IEEE/CVF Conference on Computer Vision and Pattern Recognition}, pages 14234--14243, 2020.

\bibitem[Pepa et~al.(2021)Pepa, Spalazzi, Capecci, and Ceravolo]{pepa2021automatic}
Lucia Pepa, Luca Spalazzi, Marianna Capecci, and Maria~Gabriella Ceravolo.
\newblock Automatic emotion recognition in clinical scenario: a systematic review of methods.
\newblock \emph{IEEE Transactions on Affective Computing}, 14\penalty0 (2):\penalty0 1675--1695, 2021.

\bibitem[Radford et~al.(2021)Radford, Kim, Hallacy, Ramesh, Goh, Agarwal, Sastry, Askell, Mishkin, Clark, et~al.]{radford2021learning}
Alec Radford, Jong~Wook Kim, Chris Hallacy, Aditya Ramesh, Gabriel Goh, Sandhini Agarwal, Girish Sastry, Amanda Askell, Pamela Mishkin, Jack Clark, et~al.
\newblock Learning transferable visual models from natural language supervision.
\newblock In \emph{International conference on machine learning}, pages 8748--8763. PMLR, 2021.

\bibitem[Touvron et~al.(2023)Touvron, Lavril, Izacard, Martinet, Lachaux, Lacroix, Rozi{\`e}re, Goyal, Hambro, Azhar, et~al.]{touvron2023llama}
Hugo Touvron, Thibaut Lavril, Gautier Izacard, Xavier Martinet, Marie-Anne Lachaux, Timoth{\'e}e Lacroix, Baptiste Rozi{\`e}re, Naman Goyal, Eric Hambro, Faisal Azhar, et~al.
\newblock Llama: Open and efficient foundation language models.
\newblock \emph{arXiv preprint arXiv:2302.13971}, 2023.

\bibitem[Xenos et~al.(2024)Xenos, Foteinopoulou, Ntinou, Patras, and Tzimiropoulos]{xenos2024vllms}
Alexandros Xenos, Niki~Maria Foteinopoulou, Ioanna Ntinou, Ioannis Patras, and Georgios Tzimiropoulos.
\newblock Vllms provide better context for emotion understanding through common sense reasoning.
\newblock \emph{arXiv preprint arXiv:2404.07078}, 2024.

\bibitem[Yang et~al.(2022)Yang, Huang, Wang, Liu, Zhai, Su, Li, and Zhang]{yang2022emotion}
Dingkang Yang, Shuai Huang, Shunli Wang, Yang Liu, Peng Zhai, Liuzhen Su, Mingcheng Li, and Lihua Zhang.
\newblock Emotion recognition for multiple context awareness.
\newblock In \emph{European conference on computer vision}, pages 144--162. Springer, 2022.

\bibitem[Yang et~al.(2023{\natexlab{a}})Yang, Chen, Wang, Wang, Li, Liu, Zhao, Huang, Dong, Zhai, et~al.]{yang2023context}
Dingkang Yang, Zhaoyu Chen, Yuzheng Wang, Shunli Wang, Mingcheng Li, Siao Liu, Xiao Zhao, Shuai Huang, Zhiyan Dong, Peng Zhai, et~al.
\newblock Context de-confounded emotion recognition.
\newblock In \emph{Proceedings of the IEEE/CVF Conference on Computer Vision and Pattern Recognition}, pages 19005--19015, 2023{\natexlab{a}}.

\bibitem[Yang et~al.(2023{\natexlab{b}})Yang, Huang, Xu, Li, Wang, Li, Wang, Liu, Yang, Chen, et~al.]{yang2023aide}
Dingkang Yang, Shuai Huang, Zhi Xu, Zhenpeng Li, Shunli Wang, Mingcheng Li, Yuzheng Wang, Yang Liu, Kun Yang, Zhaoyu Chen, et~al.
\newblock Aide: A vision-driven multi-view, multi-modal, multi-tasking dataset for assistive driving perception.
\newblock In \emph{Proceedings of the IEEE/CVF International Conference on Computer Vision}, pages 20459--20470, 2023{\natexlab{b}}.

\bibitem[Yang et~al.(2024)Yang, Yang, Li, Wang, Wang, and Zhang]{yang2024robust}
Dingkang Yang, Kun Yang, Mingcheng Li, Shunli Wang, Shuaibing Wang, and Lihua Zhang.
\newblock Robust emotion recognition in context debiasing.
\newblock In \emph{Proceedings of the IEEE/CVF Conference on Computer Vision and Pattern Recognition}, pages 12447--12457, 2024.

\bibitem[Zhang et~al.(2019)Zhang, Liang, and Ma]{zhang2019context}
Minghui Zhang, Yumeng Liang, and Huadong Ma.
\newblock Context-aware affective graph reasoning for emotion recognition.
\newblock In \emph{2019 IEEE International Conference on Multimedia and Expo (ICME)}, pages 151--156. IEEE, 2019.

\bibitem[Zhang et~al.(2023)Zhang, Wang, Tiwari, Li, Wang, and Qin]{zhang2023dialoguellm}
Yazhou Zhang, Mengyao Wang, Prayag Tiwari, Qiuchi Li, Benyou Wang, and Jing Qin.
\newblock Dialoguellm: Context and emotion knowledge-tuned llama models for emotion recognition in conversations.
\newblock \emph{arXiv preprint arXiv:2310.11374}, 2023.

\bibitem[Zhang et~al.(2024)Zhang, Zhou, and Liu]{zhang2024makes}
Yuanhan Zhang, Kaiyang Zhou, and Ziwei Liu.
\newblock What makes good examples for visual in-context learning?
\newblock \emph{Advances in Neural Information Processing Systems}, 36, 2024.

\bibitem[Zhao and Patras(2023)]{zhao2023prompting}
Zengqun Zhao and Ioannis Patras.
\newblock Prompting visual-language models for dynamic facial expression recognition.
\newblock \emph{arXiv preprint arXiv:2308.13382}, 2023.

\end{thebibliography}



\end{document}